\documentclass[11pt]{article}
\usepackage{coling2020}
\usepackage{times}
\usepackage{url}
\usepackage{latexsym}
\usepackage{amsmath}
\usepackage{graphicx}
\usepackage{adjustbox}
\usepackage{multirow}
\usepackage{subcaption}

\colingfinalcopy 

\title{Re-framing Incremental Deep Language Models for Dialogue Processing with Multi-task Learning}

\author{Morteza Rohanian, Julian Hough\\

 Cognitive Science Group\\School of Electronic Engineering and Computer Science\\
  Queen Mary University of London\\
  {\tt  \{m.rohanian, j.hough\} @qmul.ac.uk}}

\date{}

\begin{document}
\maketitle
\begin{abstract}

We present a multi-task learning framework to enable the training of one universal incremental dialogue processing model with four tasks of disfluency detection, language modelling, part-of-speech tagging, and utterance segmentation in a simple deep recurrent setting. We show that these tasks provide positive inductive biases to each other with the optimal contribution of each one relying on the severity of the noise from the task. Our live multi-task model outperforms similar individual tasks, delivers competitive performance, and is beneficial for future use in conversational agents in psychiatric treatment.

\end{abstract}

\section{Introduction}
\label{intro}

%
%
\blfootnote{
    %
    %
    %
    
    \hspace{-0.65cm}  
     This work is licensed under a Creative Commons 
     Attribution 4.0 International Licence.
     Licence details:
     \url{http://creativecommons.org/licenses/by/4.0/}.
    %
    %
}

Conversational technologies offer a remarkable addition to the current approaches for providing mental healthcare. Communications with these conversational agents have been found to include discoverable psychological distress signs, such as the rate of filled pauses, speech rate, and various temporal and turn-related characteristics \cite{gratch2014distress}. In the human-human automatic analysis of patient-doctor conversations, it has also been found that different types of disfluency can indicate levels of adherence to medication \cite{howes2012helping}. Markers of disfluency also hold predictive power for the identification of cognitive disorders \cite{Rohanian2020}.

Such devices are mainly used for processing content, which is then analyzed offline. There is much work on detecting disfluencies for offline analysis of transcripts with gold standard utterance segmentation within much of the current effort on disfluency detection on telephone conversations begun by \newcite{charniak2001edit}. However, given that these models do not operate for live systems and rely on rich transcription data, including the pre-segmentation of dialogue acts, to facilitate more cost-effective study of other data, it would be easier to be able to perform directly and incrementally off the speech signal, or at least from automatic speech recognition (ASR) results as they arrive into the system. The incremental model must work with minimum latency as it receives word-by-word data and does so without modifying its initial assumptions and providing its best decisions as soon as possible in line with the principles set out in \cite{hough2014strongly}.

We combine incremental identification of disfluencies with three other essential tasks for active conversational models to provide a favorable inductive bias to disfluency detection and to study the way these tasks interact. We explore a multi-task learning (MTL) framework to enable the training of one universal model with four tasks of disfluency detection, language modelling, part-of-speech (POS) tagging, and utterance segmentation, which in the data we use is also equivalent to dialogue act segmentation. Multi-task learning seeks to improve learning efficiency and predictive power by learning from a shared representation with multiple objectives. We investigate the entire power set of these tasks to investigate the interaction between them. We experiment with two different losses: a naive weighted sum of losses where the weights of loss are uniform and a loss function based on maximizing the Gaussian likelihood with task-dependent uncertainty.

We train and test a simple neural model for the different tasks, experiment with all the combinations of the tasks, different loss functions for each of the tasks, and experiment with different input representations (words vs. words with word duration).

\section{Related Work}

Although significant research has been done on disfluency detection as an individual task, most of this work uses transcripts as texts rather than using data from live speech, with the intention of `cleaning' such texts of disfluent content for eventual post-processing. They are performed on pre-segmented utterances in the Switchboard corpus of telephone conversations \cite{godfrey1992switchboard}. Sequence tagging models with start-inside-outside (BIO) style tags have been used in many studies to detect disfluencies. The most common techniques have utilized discriminative models as a classifier like Conditional Random Fields (CRFs) \cite{lafferty2001conditional}.

Such methods are insufficient if we intend to measure context from repairs and edit words for disfluency detection, which is useful in the psychiatric domain and logical for a dialogue framework that aims to measure a clear understanding of user statements.

Methods focusing on incremental performance have been uncommon. \newcite{hough2014strongly} used a pipeline of classifiers and language model features in a highly incrementally operating system without looking-ahead. Incremental dependency parsing paired with removal of disfluency was also investigated \cite{rasooli-tetrault-2015}. Two studies have applied recurrent neural networks for live disfluency detection. One approach, using a simple Elman Recurrent Neural Network (RNN), investigates incremental detection, with an objective coupling detection efficiency with low latency \cite{hough2015recurrent}. A development of this work proposed a joint task of incremental disfluency detection and utterance segmentation, with the performance of both tasks jointly improving over equivalent systems doing the individual tasks \cite{hough2017joint}.

There has been research on utterance segmentation as an individual task and also in joint models. \newcite{cuendet2006model} makes use of a range of lexical and acoustic features. \newcite{xu2014deep} used prosodic and lexical features to implement a DNN combined with a CRF classifier for broadcast news speech. \newcite{atterer2008towards} used syntactic ground-truth information, in a rare incremental approach, to predict whether the current word on Switchboard is the end of the utterance (dialog act). \newcite{seeker2016train} used utterance segmentation in a joint framework with dependency parsing. 

Language models have been used as an auxiliary task for disfluency detection, based on the intuition that disfluency occurrences will be indicated and edited from the context to improve the prediction of the next words \cite{Johnson-Charniak2004}, most recently using variants of RNN language models \cite{shalyminovmulti2018}. POS tags have also been used as an input feature in detecting disfluencies, with slight boosts in disfluency detection possible \cite{hough2015recurrent,hough2017joint}.

In this paper, we define a live setting of joint tasks, which include disfluency detection, utterance segmentation, language modelling, and POS tagging. After defining the tasks in the next section, we present a simple deep learning system, which simultaneously detects disfluencies, assigns POS tags, predicts upcoming utterance boundaries, and following words from incremental word hypotheses and derived information. 

\section{The Tasks}

\textbf{Incremental disfluency detection} Disfluencies are typically assumed to have a reparandum-interregnum-repair structure, in their fullest form as speech repairs \cite{Shriberg1994,Meteer-EtAl1995}. A reparandum is a stretch of speech subsequently fixed by the speaker; the corrected expression is a repair, the start of which we will refer to as the \textit{repair onset}. An interregnum word is a filler or a reference expression between the words of repair and reparandum, often a halting step as the speaker produces the repair, giving the structure as in (\ref{eg:RepairAnnotationStructure})

\vspace*{-0.3cm}
\begin{footnotesize}
\begin{equation}\label{eg:RepairAnnotationStructure}
\strut \texttt{ John }
\underbrace{\strut \texttt{ [ likes } }_\texttt{reparandum}+
\underbrace{\strut \texttt{ \{~uh~\}} }_\texttt{interregnum}
\underbrace{\strut \texttt{  loves ]} }_\texttt{repair}
\strut \texttt{ Mary}
\end{equation}
\end{footnotesize}
\vspace*{-0.2cm}

In the absence of reparandum and repair, the disfluency reduces to an isolated \textit{edit term}. A marked, lexicalized edit term such as a filled pause (``uh'' or ``um'') or more phrasal terms like ``I mean" and ``you know" can occur. Recognizing these elements and their structure is then the task of disfluency detection.

The task of detecting incremental disfluencies adds to the challenge of doing this in real time, word-by-word from left to right. Disfluency identification is then cast as the same challenge a human processor faces with a disfluent utterance: only when the interregnum is detected, or perhaps even when the repair onset is encountered, does it become apparent the earlier content is now to be regarded as ``to be repaired", i.e., identified as the reparandum. Therefore, the task cannot be established as a simple sequence labeling task in which the tags for the reparandum, interregnum, and repair phases are allocated left-to-right over words as seen in the above example; in this case, it will require the assumption that ``likes" would be repaired, at a point where there is no evidence to make it available. 

We follow \newcite{hough2015recurrent}, using a tag set that encodes the start of the reparandum only at a time when it can be inferred, mainly when the repair begins -- the individual disfluency detection task is to tag words as in the top line of tags in Fig.~\ref{fig:bostondenver} as either fluent ($f$) an edit term ($e$), a repair onset word ($rpS{-}N$ for the reparandum starting $N$ words back) and a repair end word of the type repeat ($rpnRep$), substitution ($rpnSub$) or delete ($rpnDel$).

\textbf{Incremental utterance segmentation} Utterance segmentation has a strong interdependence with disfluency detection, as standard disfluency detection operates on an utterance or dialogue act level. The two tasks may be more difficult done separately. Without utterance segmentation, disfluent restarts and repairs may be predicted at fluent utterance boundaries \cite{hough2017joint}, while without disfluency detection, the utterance segmentation may predict an utterance boundary at a repair onset point. So we presume that the tasks can be carried out together in a supplementary manner.

\begin{figure*}[t]
\begin{small}
\centering
\setlength\tabcolsep{0em} 
\begin{tabular}{p{3.3cm}lllcccclll}
&$|$ A~~uh~~flight~&~~[~~to Boston & ~+~\{ uh & ~I & ~mean \} & ~to & ~Denver~] & ~on Friday $|$ & ~~~~~~~& Thank you  $|$\\
Disfluency &~~$\scriptstyle f$~~~~~$\scriptstyle e$~~~~~$\scriptstyle f$~~&~~~~~~~$\scriptstyle f$~~~~$\scriptstyle f$~ & ~~~~~~~~$\scriptstyle e$~~~ & ~~~$\scriptstyle e$~~~ & ~~$\scriptstyle e$~~~  & ~~~~$\scriptstyle rpS-5$~~~ & ~~$\scriptstyle rpnSub$~~~~ & ~~~$\scriptstyle f$~~~ $\scriptstyle f$~~~ & ~~~~~~~& ~~~~$\scriptstyle f$~~~~~$\scriptstyle f$  \\
Utterance segmentation &~.$\scriptstyle w$-~~~-$\scriptstyle w$-~~-$\scriptstyle w$-~&~~~~~~-$\scriptstyle w$- -$\scriptstyle w$- & ~~~~~ -$\scriptstyle w$-~~ & ~-$\scriptstyle w$-~~ & ~-$\scriptstyle w$-~~  & ~~~-$\scriptstyle w$-~~~~ & ~-$\scriptstyle w$-~~~ & ~-$\scriptstyle w$-~~ -$\scriptstyle w$.~~ & ~~~~~~~& ~~~.$\scriptstyle w$-~~~-$\scriptstyle w$.  
\\
POS tags &$\scriptstyle DT$~~~$\scriptstyle UH$~~$\scriptstyle NN$~&~~~~~~$\scriptstyle IN$ $\scriptstyle NNP$ & ~~~~~ $\scriptstyle UH$~~ & ~$\scriptstyle PRP$~~ & ~$\scriptstyle VB$~~  & ~~~$\scriptstyle IN$~~ & ~$\scriptstyle NNP$~~~ & ~$\scriptstyle IN$~~ $\scriptstyle NNP$~~ & ~~~~~~~& ~~~$\scriptstyle VB$~~~$\scriptstyle PRP$ 
\\
\end{tabular}
\end{small}
\caption{An utterance with the disfluency tags (repair onsets and edit terms), utterance segmentation annotation tags and POS tags in our incremental tag schemes}\label{fig:bostondenver}
\end{figure*}

We characterize incremental utterance segmentation as the real-time word-by-word decision as to whether the current utterance is ending. We are shifting from a strictly reactive, silent-signaled approach to prediction. Following \newcite{hough2017joint}, we use four tags to define ranges of acoustic data, which are the time spans of forced aligned gold standard words, equivalent to a BIES utterance scheme (Beginning, Inside, End and Single) to allow for prediction. The tagset allows information to be captured from the previous context of the word to determine whether this word continues an existing utterance (the \texttt{-} prefix) or begins anew (the \texttt{.} prefix), and also allows online prediction of whether the following word will continue the existing utterance (the \texttt{-} suffix) or whether the current word completes the utterance (the \texttt{.} suffix)- see Fig.~\ref{fig:bostondenver} for the incremental utterance segmentation tags around a $w$ symbol for the current word. Unlike \newcite{hough2017joint}, we do not employ a joint tag set, instead looking to use multi-task learning as described below to combine these tasks into one learning regime.

\textbf{Incremental POS tagging} Part-of-speech (POS) tags can improve disfluency detection on different settings. We combine POS tagging as an additional task with a similar structure to utterance segmentation and disfluency detection to motivate the model to learn better features for syntactic and semantic structures, which can improve the other tasks without requiring additional training data. POS tagging could be helped by information about utterance segmentation and disfluency structure as the parallelism between the reparandum and repair in substitutions, as shown in the repeated $IN~NNP$ sequences in Fig.~\ref{fig:bostondenver} show, allowing better disambiguation of `to' with this extra information.

\textbf{Language modelling} The central idea of our approach to language modelling is that the probability of the current word can be more accurately modelled with knowledge from the other dialogue phenomena detected as described above, compared to the standard approach of just using the previous word values, modelling it as part of a joint task \cite{heeman1999speech}. A secondary assumption is that disfluencies, each of which has a context-conditioned probability, can be represented as word-like occurrences themselves \cite{stolcke1996statistical}. Utterance segmentation and disfluency detection are optimized only on the basis of accurate labels. While each token in the input has a target tag, some contribute next to nothing to the training process. The disfluency detection and utterance segmentation models can acquire a bias in label distribution by gaining additional information from the majority labels. So we are using another objective that would allow the models to make full use of the dataset.

In addition to learning how to predict each word's labels, we also suggest optimizing different parts of the architecture as a language model. The role of predicting the next word would require the model to study more general patterns of semantic and syntactic structure, which can then be reproduced to predict individual words more accurately. This purpose is also generalizable to every task and dataset that does not include additional annotated training details. With a simple adjustment in the model, we introduce a second parallel output layer for each token, improving it to predict the next word. We predict the next word in the sequence only based on the hidden representation from the left-to-right moving LSTM.

\subsection{Combining the tasks with multi-task learning}
In this paper, following the intuition of \newcite{heeman1999speech} that combining these tasks will improve the accuracy of all of them, given their mutual dependence, we take a multi-task learning (MTL) deep learning setting, where we have several prediction tasks over the same input space. Disfluency detection, utterance segmentation, language modelling, and POS-tagging are all done simultaneously at each input step. The method can learn how to efficiently adjust the task weightings, resulting in a better performance than individual learning of each task. The key to optimization is the different type of loss function used for each task, which we investigate in our experiments below.

\section{Model architecture}

\begin{figure}[!t]
\centering

\includegraphics[height=8cm]{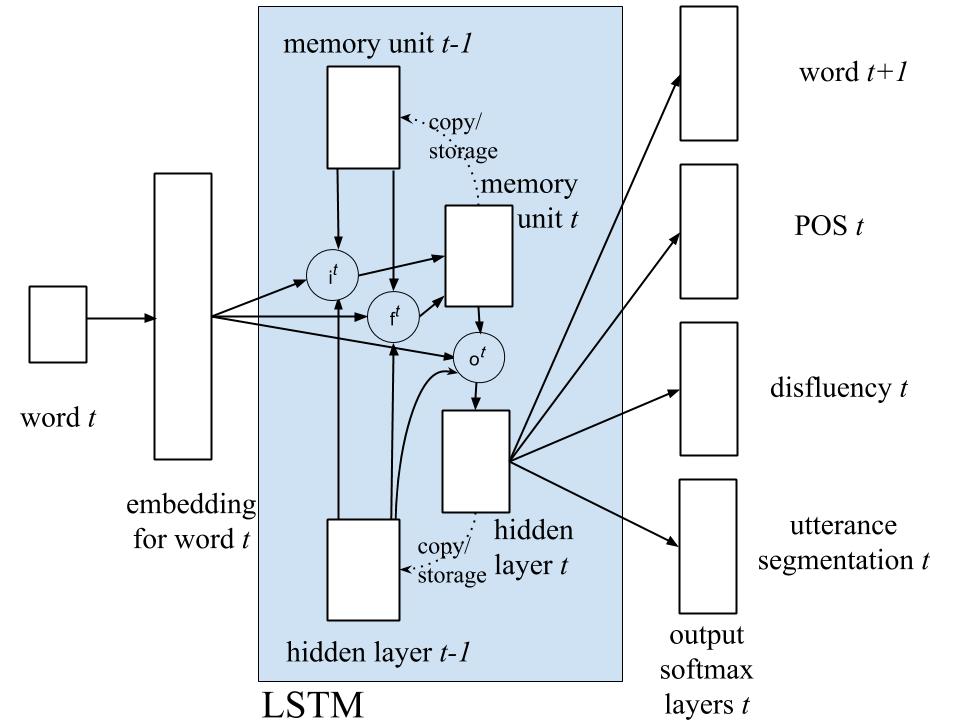}
\caption{Incremental Long Short-Term Memory net (LSTM) for disfluency detection, utterance segmentation, POS-tagging tasks and language modelling.}\label{fig:network}
\end{figure}

For this task, we use the incremental neural network model the Long Short-Term Memory net (LSTM), as the baseline architecture. For our individual disfluency detection, utterance segmentation, and POS-tagging tasks, the model takes the input and uses an LSTM to add a label to each token. For our language modelling task, the model predicts the likelihood of the current word given the previous words. The overall architecture of our model is shown in Figure 2. First, the input tokens are mapped to distributed word embeddings. An LSTM is used for constructing context-dependent representations for every word \cite{rei2017semi}. The LSTM, together with the word embedding from the current step, takes the hidden state from the preceding time step as input and produces a new hidden state:

\begin{equation}
h_t = LSTM (x_t, h_{t-1})   
\end{equation}

\noindent
Then, the representation is passed through a feedforward layer:

\begin{equation}
d_t = tanh (W_dh_t)   
\end{equation}
\noindent
where $tanh$ is a non-linear activation function and $W_d$ is weight matrix.

We use a Conditional Random Field (CRF) output architecture to predict a tag for each token. Although this model creates predictions according to all the words in the data, the labels are predicted individually. There are high dependencies for disfluency detection between subsequent labels, and explicit modeling of those connections can be helpful. The output can be modified to have a CRF, enabling the model to test for the most optimal path across all available label sequences. The model is then improved by maximizing the score for the correct label sequence while reducing the score for all other sequences:

\begin{equation}
E = -s(y) + log \sum _{\widetilde{y} \in \widetilde{Y}} e^{s(\widetilde{y})} 
\end{equation}
where $s(y)$ is the score for the sequence $y$ and $Y$ is the set of all tag sequences.

For our language modelling task, first, using a non-linear layer, the hidden representations from LSTMs are mapped to a new space:

\begin{equation}
m_t = tanh (W_m h_t)   
\end{equation}
where $W_m$ is weight matrix. This distinct transformation learns to obtain features that are specific to language modeling. Then, these representations are carried through softmax layers to predict the likelihood of the current word $w_t$ given the previous words characterized by $m_t$:

\begin{equation}
P(w_{t}\mid m_t) = softmax(W_q m_{t-1})
\end{equation}

The objective function is an objective of regular language modeling $E_l$, which is the negative log-likelihood of the current word in the sequence given the previous words.

\begin{equation}
E_l = - log (P(w_{t}\mid m_{t}))
\end{equation}

The network is designed to predict the current label at each token position in terms of disfluency detection, utterance segmentation, and POS tagging and the likelihood of the next word in sequence by predicting the likelihood of each word in the vocabulary. The added language modeling objective enables the model to learn better representations of words that are then used for other tasks. To boost MTL's performance on the target dataset and to learn several tasks simultaneously, we propose two loss function methods: first, combining multi-objective losses with a naive approach and simply carrying out a weighted linear loss sum for each task:

\begin{equation}
\widetilde{E} = E + \alpha(E_{l})
\end{equation}

$\alpha$ is the parameter that we use to control the importance of auxiliary tasks (language modelling here). Secondly, inspired by \cite{kendall2018multi}, we use an orthogonal approach, taking the uncertainty of each task into account. This multi-task loss learns various classification and regression losses of varying quantities and units simultaneously using task uncertainty. We adjust the relative weight of each task in cost function by deriving a multi-task loss function based on maximizing the Gaussian likelihood with task-dependent uncertainty, generalizing this to our four tasks, each with their separate $\sigma$ weight:

\begin{equation}
\widetilde{E} = \sum _{i=1} ^ 4 \frac{E_i}{\sigma_i^2} + \sum _{i=1} ^ 4 log(\sigma_i)
\end{equation}

with $\sigma$ the model’s trainable observation noise parameter – capturing how much noise we have in the outputs. $E_i$ is one of four indexed to the individual tasks: $E_{LM}$, $E_{Disf}$, $E_{seg}$ and $E_{POS}$.

\section{Experimental Set-up}

With various tasks, we assess the proposed architecture in joint models with a different number of tasks. We experimented with two types of inputs: word embeddings and the duration of the current word in addition to the word embeddings. The word embedding was initialized with pre-trained vectors available to the public, generated using 50-dimensional embedding trained on Google News \cite{mikolov2013distributed}. The neural network model has been implemented using Tensorflow \cite{abadi2016tensorflow}. The LSTM hidden layers are 200. To minimize computational complexity in these experiments, the language modeling task predicted only the 7000 most commonly used words, with an additional token representing all the other words. Results on the development set were also used to find the best model to be evaluated on the test set. We train all models for a maximum of 50 epochs; otherwise, stop training if there is no advancement on the best score on the transcript validation set after 7 epochs. The code used in the experiments is publicly available in an online repository.\footnote{https://github.com/mortezaro/mtl-disfluency-detection}

\textbf{Data} We use standard Switchboard training data (all conversation numbers starting sw2*,sw3 * in the Penn Treebank III release: 100k utterances, 650 K words) and use standard held-out data (PTB III files sw4[5-9] *: 6.4 K utterances, 49 K words) as our validation set. We test standard test data (PTB III files 4[0-1] *) with partial words and punctuation stripped away from all files.

\textbf{Evaluation Criteria} We calculate F1 accuracy for repair onset detection $F_{rpS}$ and for edit term words $F_{e}$, which includes interregna. For utterance segmentation we also use word-level F1 scores for utterance boundaries (end-of- utterance words) $F_{uttSeg}$. We calculate accuracy $ACC_{POS}$ for POS tagging and $Perplexity$ for the language modelling task.

We measure latency and the stability of output over time, which is essential to the model's live nature. We use the first time to detection (FTD) metric of \cite{zwarts2010detecting} for latency: the average distance (in the number of words) consumed before the first detection of gold standard repairs from the repairs onset word. For stability, from the evaluation of incremental processors by \cite{baumann2011evaluation}, we evaluate the edit overhead (EO) of the output labels –  the proportion of the unnecessary editing (insertion and deletion) necessary to attain the final labels produced by the model.

\begin{table}[!t]
\caption{Non-incremental (dialogue-final) results for different tasks with a loss function based on uncertainty of tasks}
\label{tab:results-test}
\centering
\begin{adjustbox}{width=1\textwidth}
\begin{tabular}{|clccccc|}
\hline
\multicolumn{1}{|c|}{Input}                            & \multicolumn{1}{c|}{Models}    & \multicolumn{1}{c|}{$F_{uttSeg}$} & \multicolumn{1}{c|}{$F_{e}$} & \multicolumn{1}{c|}{$F_{rpS}$} & \multicolumn{1}{c|}{$ACC_{POS}$} & $Perplexity$  \\ \hline
                                                       & \multicolumn{6}{c|}{\textbf{Baselines}}                                                                                                                                               \\ \hline
\multicolumn{1}{|c|}{Words}                            & Seeker et al. (2016)           & 0.767                             & -                            & -                              & -                                & -             \\ \cline{1-1}
\multicolumn{1}{|c|}{Words}                            & Hough and Schlangen (2015)     & -                                 & 0.902                        & 0.689                          & -                                & -             \\ \cline{1-1}
\multicolumn{1}{|c|}{Words / Words + Timings}          & Hough and Schlangen (2017)     & 0.748                             & 0.918                        & 0.720                          & -                                & -             \\ \hline
                                                       & \multicolumn{6}{c|}{\textbf{Our Models}}                                                                                                                                              \\ \hline
\multicolumn{1}{|c|}{\multirow{7}{*}{Words}}           & Single Tasks                   & 0.689                             & 0.904                        & 0.678                          & 0.961                            & 65.3          \\
\multicolumn{1}{|c|}{}                                 & LM + uttSeg                    & 0.725                             & -                            & -                              & -                                & 65.0          \\
\multicolumn{1}{|c|}{}                                 & LM + Disf                      & -                                 & 0.915                        & 0.717                          & -                                & 67.5          \\
\multicolumn{1}{|c|}{}                                 & POS + Disf                     & -                                 & 0.913                        & 0.713                          & 0.961                            & -             \\
\multicolumn{1}{|c|}{}                                 & uttSeg + Disf                  & 0.709                             & 0.917                        & 0.718                          & -                                & -             \\
\multicolumn{1}{|c|}{}                                 & uttSeg + Disf + LM       & 0.734                             & 0.917                        & 0.724                          & \iffalse0.963\fi-                            & \textbf{64.3} \\
\multicolumn{1}{|c|}{}                                 & uttSeg + Disf + LM + POS & 0.763                             & 0.917                        & \textbf{0.743}                 & \textbf{0.965}                   & \textbf{64.3} \\ \hline
\multicolumn{1}{|c|}{\multirow{6}{*}{Words + Timings}} & LM + uttSeg                    & 0.728                             & -                            & -                              & -                                & 65.2          \\
\multicolumn{1}{|c|}{}                                 & LM + Disf                      & -                                 & 0.915                        & 0.711                          & -                                & 67.5          \\
\multicolumn{1}{|c|}{}                                 & POS + Disf                     & -                                 & 0.913                        & 0.712                          & 0.960                            & -             \\
\multicolumn{1}{|c|}{}                                 & uttSeg + Disf                  & 0.712                             & 0.917                        & 0.715                          & -                                & -             \\
\multicolumn{1}{|c|}{}                                 & uttSeg + Disf + LM       & 0.738                             & 0.919                        & 0.720                          & -                            & 64.4          \\
\multicolumn{1}{|c|}{}                                 & uttSeg + Disf + LM + POS & \textbf{0.767}                    & \textbf{0.922}               & 0.741                          & 0.964                            & 64.5          \\ \hline
\end{tabular}
\end{adjustbox}
\end{table}

\section{Results}

Our best utterance-final accuracies from different tasks with a loss function based on uncertainty are shown in Table~\ref{tab:results-test}. Our best $F_{rpS}$ reaches 0.743 and best $F_e$ reaches 0.922. For utterance segmentation,  $F_{uttSeg}$ reaches 0.767. It's difficult to compare our results with the standard approaches for the detection of disfluency as they use pre-segmented utterances. However our best result is as high as \cite{seeker2016train}’s models on the Switchboard data for utterance segmentation that is state-of-the-art. In comparison to incremental approaches, we outperform \newcite{hough2017joint}'s 0.748 on end-of-utterance and \newcite{hough2015recurrent} and \newcite{hough2017joint}'s 0.720 and 0.918 on $F_{rpS}$ and $F_e$.

The models using the timing outperform those with lexical information only on the utterance segmentation metrics and $F_e$ whilst having lower performance on $F_{rpS}$, language modelling, and POS tagging. We get our highest results in all tasks from the model with four joint tasks. Edit term detection works very well at 0.922, nearing the state-of-the-art on the Switchboard reported at 0.938. Our lowest perplexity is 64.3. Additional training objective leads to more accurate language modelling in all joint models except when it is trained with disfluency detection alone. The best result for POS tagging is 0.965. It is important to mention that it's difficult to do LM and POS tagging comparisons on this particular dataset. However, the high quality of our neural POS tagging and language modelling on un-segmented data and in a live setting are comparable to state-of-the-art approaches and useful enough to improve utterance segmentation and disfluency detection, in line with our main goal.

\textbf{Joint tasks} As can be seen in the Table~\ref{tab:results-test}, the overall highest scoring systems for individual tasks do not achieve results in any relevant metric of the top performing joint systems. We achieve the highest $F_{rpS}$, $F_e$, $F_{uttSeg}$, $ACC_{POS}$ and $Perplexity$ with a model with all four tasks. Adding language modelling, utterance segmentation, and POS tagging also help disfluency detection in models with two and three joint tasks. While the performance improvements are small, they are consistent across all joint models. Joint models with more than two tasks do not have an impact on $F_e$ when the input is only lexical. Our joint model with four tasks outperforms \newcite{hough2017joint} 's joint model in both disfluency detection and utterance segmentation.

$F_{uttSeg}$ is the metric with the most improvements with the addition of relevant tasks as it gets higher in all the joint model. Language modelling helps utterance segmentation more than disfluency detection in joint models with two tasks. POS tagging does not help utterance segmentation and disfluency detection as much as language modelling in joint models with two tasks but improves the results compared to single tasks.

We also get the best results for language modelling with the joint model with three and four tasks ($Perplexity$ 64.3). Adding disfluency detection to language modelling on its own actually decreases the performance ($Perplexity$ 65.3 vs.\ 67.5). Utterance segmentation is the single most beneficial auxiliary task for language modelling ($Perplexity$ 65.0).

The baseline performance in POS tagging is close to the upper bound; therefore the language modelling, disfluency detection, and utterance segmentation objectives do not provide much added advantage.

\textbf{Timing as input} Adding timing information as input provides a consistent improvement across all joint tasks in comparison to single tasks, as we can observe in Table~\ref{tab:results-test}. Timing improves utterance segmentation performance in models with two tasks more than other tasks. Timing can help to obtain more information for detecting edit terms in joint tasks. The largest benefit from the timing was observed on the joint task of utterance segmentation and edit term detection. Language modelling performance decreases in all joint models with word timing compared to models without it except in the joint model with disfluency detection.

\textbf{Utility of loss function using uncertainty} From Table~\ref{tab:results-naive} we observe that our `naive' loss function baseline simply using the sum of all the separate losses leads to a decrease of performance in all tasks in joint models compared to the uncertainty loss functions except language modelling. Utterance segmentation gets the most unfavorable results with naive loss function compared to other tasks. Adding disfluency detection decreases the utterance segmentation performance the most when it is an auxiliary task. Language modelling is the best auxiliary task across different joint models and gets the best results compared to the same models with loss functions with uncertainty. In some cases, using naive loss functions does not lead to an improvement in comparison to the single-task models. Single-task utterance segmentation outperforms all the joint models with naive loss except the joint model with language modelling in $F_{uttSeg}$. Single-task disfluency detection gets better $F_e$ than all the joint models with naive loss except the joint model with POS tagging. Single POS tagging gets better $ACC_{POS}$ than all joint models with naive loss. An increasing number of tasks makes it more difficult to obtain optimal weights roughly. Using the uncertainty loss function improves all tasks in all models, but utterance segmentation and disfluency detection performances improve the most.

We show that performances in multi-task learning settings for all of our tasks rely heavily on the suitable choice of loss weightings. We note that each task's optimal weighting depends on the severity of the noise from the task.

\textbf{Incrementality} As shown in Table~\ref{tab:results-incremental}, the incremental performance of our individual disfluency task was better overall than the best performance in joint model settings, with low EO in our best setting at 1.08 and FTD just slightly above 1 word (1.01). If we define EO as the proportion of unnecessary edits to get to the final labels, we can see that adding language modelling increases EO significantly and makes the system less stable (2.01). Adding other classification tasks (POS tagging and utterance segmentation) does not decrease the incremental performance as much as language modelling.

In terms of FTD, all of our models are very fast and close to 1. Our individual disfluency model gets the best score. Adding language modelling slightly decreases the incremental performance for FTD (1.09 vs.\ 1.01). Joint disfluency detection models with utterance segmentation, POS tagging, and language modelling incremental performance are still comparable to individual disfluency detection model in FTD performance. Overall, incremental disfluency tagging performance is decreased most by adding language modelling as a joint task, but this comes with the trade-off of better final accuracy.

\begin{table}[]
\caption{Non-incremental (dialogue-final) results for different tasks with a naive loss function}
\label{tab:results-naive}
\centering
\begin{tabular}{|lccccc|}
\hline
\multicolumn{1}{|c|}{Models}   & \multicolumn{1}{c|}{$F_{uttSeg}$} & \multicolumn{1}{c|}{$F_{e}$} & \multicolumn{1}{c|}{$F_{rpS}$} & \multicolumn{1}{c|}{$ACC_{POS}$} & $Perplexity$ \\ \hline
Single Tasks                   & 0.689                             & 0.904                        & 0.678                          & 0.961                            & 65.3 \\\hline
LM + uttSeg                    & 0.691                             & -                            & -                              & -                                & 64.8         \\
LM + Disf                      & -                                 & 0.903                        & 0.687                          & -                                & 65.5         \\
POS + Disf                     & 0.665                             & 0.904                        & 0.684                          & 0.957                            & -            \\
uttSeg + Disf                  & 0.683                           & 0.907                        & 0.697                          & -                                & -            \\
uttSeg + Disf + LM       & 0.667                             & 0.903                        & 0.682                          & \iffalse0.953\fi-                            & 64.5         \\
uttSeg + Disf + LM + POS & 0.662                             & 0.902                        & 0.690                          & 0.956                            & 64.5         \\ \hline
\end{tabular}
\end{table}

\section{Error Analysis}

\textbf{Different repair types} Categorizing repetitions as verbatim repeats, substitutes as the other repairs marked with a repair phase, and deletes as those without one, we see in Table~\ref{tab:results-types} that our joint models get better results than the separate disfluency detection task on repeat and substitution types. Separate disfluency detection outperforms joint models in detecting deletes. It seems that adding other tasks to joint models does not improve the accuracy of detecting the rare delete repairs but helps the accuracy of substitution repairs.

Adding other tasks to the task of disfluency detection improves the repetitions detection, but we do not observe an improvement in models with more than two tasks. While the performance improvements of joint models in detecting substitutions are small, they are consistent across all models with different tasks. We get the best result for substitutions in our single disfluency model at 0.65 and for repeats in our joint model with language modelling at 0.93 when the loss function is naive. For deletes, the single disfluency detection model and joint segmentation and disfluency detection models get the best results at 0.34.

\begin{table}[htb]
\begin{minipage}[t]{0.52\linewidth}\centering
        \caption{$F_{rpS}$ repairs with different reparandum lengths}
        \label{tab:results-length}

        \raggedright
\begin{tabular}{|p{4.1cm}p{0.35cm}p{0.35cm}p{0.35cm}p{0.35cm}p{0.4cm}|}
\hline
Models                   & ~1                      & ~2                      & ~3                      & ~4                      & ~5                       \\ \hline
Disf                     & .84                     & .67                     & .40                     & .31                     & .13                      \\
LM + Disf                & .84                     & .68                     & .42                     & .33                     & .13                      \\
POS + Disf               & .84 & .67 & .41 & .31 & .13 \\
uttSeg + Disf            & .84                     & .67                     & .40                     & .30                     & .13                      \\
uttSeg + Disf + LM       & .84                     & .68                     & .43                     & .33                     & .13                      \\
uttSeg + Disf + LM + POS & .85                     & .68                     & .43                     & .33                     & .13                      \\ \hline
\end{tabular}
\end{minipage}\hfill%
\begin{minipage}[t]{0.44\linewidth}\centering
        \raggedleft
\caption{Incremental results for disfluency detection}
\label{tab:results-incremental}
\begin{tabular}{|lll|}
\hline
Models                   & FTD & EO \\ \hline
Disf                     & 1.01                          & 1.08               \\
LM + Disf                & 1.09                          & 2.01               \\
POS + Disf               & 1.03                          & 1.14               \\
uttSeg + Disf            & 1.02                          & 1.11               \\
uttSeg + Disf + LM       & 1.10                          & 2.05               \\
uttSeg + Disf + LM + POS & 1.12                          & 2.04               \\ \hline
\end{tabular}
\end{minipage}

\end{table}

\begin{table}[!ht]
        \caption{$F 1$ for repairs with different types}
\label{tab:results-types}
\centering

\begin{tabular}{|lcccccc|}
\hline
\multicolumn{1}{|l|}{}       & \multicolumn{3}{c|}{Uncertainty-based loss} & \multicolumn{3}{c|}{Naive loss}  \\ \hline
\multicolumn{1}{|l|}{Models} & Repeats     & Substitution     & Deletes    & Repeats & Substitution & Deletes \\ \hline
Disf                         & .94        & .70             & .48       & .91    & .65         & .34    \\
LM + Disf                    & .96        & .71             & .46       & .93    & .63         & .32    \\
POS + Disf                   & .96        & .70             & .47       & .90    & .63         & .33    \\
uttSeg + Disf                & .96        & .70             & .46       & .90    & .64         & .34    \\
uttSeg + Disf + LM           & .96        & .71             & .46       & .91    & .63         & .31    \\
uttSeg + Disf + LM + POS     & .96        & .72             & .46       & .91    & .63         & .30    \\ \hline
\end{tabular}
\end{table}

\textbf{Different repair lengths} While our best system still suffers from a vanishing gradient problem to predict repairs with longer reparanda,  we can observe in Table~\ref{tab:results-length} an F1 boost for repair lengths 1 to 4 in our joint models. Our joint model with four tasks gets the best results with repair length 1 to 4. 

Our joint models do not show improvements in detecting longer repairs with reparanda of 5 words or longer. The combined number of instances in the test set with reparanda over 5 words is 59, and our best model predicts 26 of them correctly.

\section{Conclusion}

We have presented a multi-task learning framework to enable the training of one universal incremental model with four tasks of disfluency detection, language modelling, part-of-speech tagging and utterance segmentation. We have observed that these tasks produce favorable inductive biases to each other, with utterance segmentation and disfluency detection getting the most benefits. We note that each task's optimal weighting relies heavily on the severity of the noise from the task. We showed that word timing information helps utterance segmentation and disfluency detection in an online setting, and adding new tasks with the exception of language modelling does not have a remarkable negative effect on the incremental metrics. 

The results show that our framework can be suitable for online conversational systems, such as conversational agents in the mental health domain. In future work, we intend to analyze the interactions between different tasks as they occur in real time. Monitoring the interaction after each word could help highlight informative moments that contribute more to optimisation of our models. Furthermore, we intend to use raw acoustic features as the input for a strongly time-linear model.

\section*{Acknowledgments}
We thank the anonymous COLING reviewers for their helpful comments and Matthew Purver for his continuous support and supervision on the wider project.

\end{document}